\documentclass{article}
\usepackage{style/foci}
\usepackage{tgpagella}
\usepackage{tgheros}
\usepackage[utf8]{inputenc}
\usepackage[T1]{fontenc}
\usepackage{microtype}

\usepackage{amsmath}
\usepackage{amssymb}
\usepackage{amsfonts}
\usepackage{amsthm}
\usepackage{mathtools}
\usepackage{mathrsfs}
\usepackage{physics}
\usepackage{braket}
\usepackage{slashed}
\usepackage{nicefrac}
\usepackage{textcomp}
\usepackage{dsfont}
\usepackage{bbm}
\usepackage{bm}

\usepackage{graphicx}
\usepackage{subcaption}
\usepackage[export]{adjustbox}
\usepackage{float}
\usepackage{booktabs}
\usepackage{dcolumn}
\newcolumntype{d}[1]{D{.}{.}{#1}}
\usepackage{bigstrut, tabularx, multirow, makecell, diagbox}
\usepackage{colortbl}
\usepackage{tabularray}
\UseTblrLibrary{booktabs}
\usepackage{threeparttable}
\usepackage{tablefootnote}
\usepackage{fontawesome5}

\usepackage{placeins}
\usepackage{caption}
\usepackage{footnote}
\usepackage{enumitem}
\usepackage{multicol}
\usepackage{xspace}
\usepackage{titletoc}
\usepackage{titlesec}
\usepackage[bottom]{footmisc}
\usepackage{setspace}

\usepackage{wrapfig}
\usepackage{tikz}
\usepackage{quantikz}
\usepackage{dashbox}
\usepackage{mdframed}
\usepackage{marvosym}
\usepackage{pifont}
\usepackage{CJK}
\usepackage{url}

\usepackage[table,x11names]{xcolor}
\usepackage[most]{tcolorbox}
\tcbuselibrary{breakable}
\usetikzlibrary{decorations.pathreplacing, fit}

\definecolor{primaryblue}{HTML}{0066CC}
\definecolor{accentcyan}{HTML}{00D4AA}
\definecolor{warmorange}{HTML}{FF6B35}
\definecolor{deepgray}{HTML}{2C3E50}
\definecolor{lightgray}{HTML}{F8F9FA}
\definecolor{gradientstart}{HTML}{667eea}
\definecolor{gradientend}{HTML}{764ba2}

\definecolor{citecolor}{HTML}{0071bc}
\definecolor{citeblue}{RGB}{0, 113, 188}
\definecolor{linkcolor}{HTML}{9A4D92}
\definecolor{firebrick}{rgb}{0.698,0.133,0.133}

\definecolor{paleviolet}{HTML}{E1EEFC}
\definecolor{CarolinaUltraLight}{HTML}{E7F4FC}
\definecolor{lightgrey}{RGB}{247, 247, 247}
\definecolor{shadecolor}{HTML}{EFEFEF}
\definecolor{lightyellow}{rgb}{1.0, 0.95, 0.7}
\definecolor{lightblue}{rgb}{0.90, 0.95, 1.0}
\definecolor{light-gray}{gray}{0.95}

\definecolor{darkgrey}{rgb}{0.5, 0.5, 0.5}
\definecolor{darkgreen}{rgb}{0, 0.5, 0}
\definecolor{mydarkblue}{rgb}{0,0.08,0.45}
\definecolor{mydarkblue2}{rgb}{0.133, 0.133, 0.698}
\definecolor{echodrk}{HTML}{0099cc}
\definecolor{mymauve}{rgb}{0.58,0,0.82}
\definecolor{midnightblue}{rgb}{0.1,0.1,0.44}
\definecolor{oxfordblue}{rgb}{0.0,0.13,0.28}
\definecolor{prussianblue}{rgb}{0.0,0.19,0.33}
\definecolor{coolteal}{rgb}{0, 0.45, 0.45}
\definecolor{olive}{rgb}{0.1, 0.3, 0}
\definecolor{mypurple}{rgb}{0.5,0,0.5}
\definecolor{almond}{rgb}{0.94, 0.87, 0.8}

\definecolor{blue_ampEncoding}{HTML}{DAE8FC}
\definecolor{green_encoder}{HTML}{D5E8D4}
\definecolor{purple_decoder}{HTML}{E1D5E7}
\definecolor{yellow_measure}{HTML}{FFF2CC}
\definecolor{gray_block}{HTML}{F5F5F5}
\definecolor{pink_dru}{HTML}{FAD9D5}
\definecolor{orange_v}{HTML}{FAD7AC}

\definecolor{colorA}{rgb}{1,0,0}
\definecolor{colorB}{rgb}{0,0.3,1}
\definecolor{colorC}{rgb}{0.9,0.8,0.2}
\definecolor{colorD}{rgb}{0,0.65,0}
\definecolor{lesslightgray}{rgb}{0.5,0.5,0.5}
\definecolor{fundamental}{RGB}{55, 110, 111}
\definecolor{Gred}{RGB}{219, 50, 54}
\definecolor{ToCgreen}{RGB}{0, 128, 0}
\definecolor{Sepia}{RGB}{112, 66, 20}
\definecolor{Dblue}{rgb}{0,0.08,0.45}
\definecolor{Blue}{rgb}{0, 0, 0.8}
\definecolor{blue}{rgb}{0,0,1}
\definecolor{UNCblue!10}{rgb}{0.84,0.91,0.98}
\definecolor{RowAlt}{rgb}{0.98,0.98,0.99}

\definecolor{CarolinaBlue}{HTML}{7BAFD4}        % Official UNC Carolina Blue
\definecolor{CarolinaLightBlue}{HTML}{B3D4E5}   % Lighter version
\definecolor{CarolinaUltraLight}{HTML}{E8F4F8}  % Very light for background
\definecolor{CarolinaText}{HTML}{1C2B33}        % Dark text color

\usepackage[pagebackref=true,breaklinks=true,colorlinks,hyperfootnotes=false]{hyperref}
\hypersetup{
  colorlinks,
  citecolor=citeblue,
  linkcolor=firebrick,
  urlcolor=firebrick
}
\usepackage[nameinlink,capitalize,noabbrev]{cleveref}

\titlespacing\section{0pt}{4pt plus 4pt minus 2pt}{-2pt plus 2pt minus 2pt}
\titlespacing\subsection{0pt}{2pt plus 4pt minus 2pt}{-2pt plus 2pt minus 2pt}
\titlespacing\subsubsection{0pt}{2pt plus 4pt minus 2pt}{-2pt plus 2pt minus 2pt}

\makeatletter
\def\th@remark{%
  \thm@headfont{\bfseries}%
  \normalfont % body font
  \thm@preskip\topsep \divide\thm@preskip\tw@
  \thm@postskip\thm@preskip
}
\makeatother

\theoremstyle{definition}

\tcbset{focithm/.style={
  enhanced, breakable, frame hidden,
  colback=black!5,
  width=\linewidth,
  enlarge left by=0pt, enlarge right by=0pt,
  boxsep=6pt, boxrule=0pt,
  left=4pt, right=4pt, top=1pt, bottom=1pt,
  arc=8pt,
  before skip=\topsep, after skip=\topsep
}}

\tcolorboxenvironment{theorem}{focithm}

\tcolorboxenvironment{lemma}{focithm}

\tcolorboxenvironment{corollary}{focithm}

\tcolorboxenvironment{proposition}{focithm}

\tcolorboxenvironment{definition}{focithm}

\tcolorboxenvironment{assumption}{focithm}

\tcolorboxenvironment{claim}{focithm}

\tcolorboxenvironment{problem}{focithm}

\tcolorboxenvironment{question}{focithm}

\newtcolorbox{titleblock}{
  enhanced,
  frame hidden,
  colback=CarolinaUltraLight,
  colframe=CarolinaUltraLight,
  boxrule=0pt,
  arc=10pt,
  left=14pt,
  right=14pt,
  top=14pt,
  bottom=14pt,
  width=\linewidth,
  before skip=12pt plus 4pt,
  after skip=12pt plus 4pt,
  grow to left by=1.5pt,
  grow to right by=1.5pt,
  before upper={
    \setlength{\parindent}{0cm}
    \setlength{\parskip}{0.5cm}
  }
}

\crefname{theorem}{Theorem}{Theorems}
\crefname{proposition}{Proposition}{Propositions}
\crefname{lemma}{Lemma}{Lemmas}
\crefname{corollary}{Corollary}{Corollaries}
\crefname{definition}{Definition}{Definitions}
\crefname{assumption}{Assumption}{Assumptions}
\crefname{remark}{Remark}{Remarks}
\crefname{problem}{Problem}{Problems}
\crefname{property}{Property}{property}
\crefname{question}{Question}{Questions}

\numberwithin{equation}{section}
\numberwithin{theorem}{section}
\numberwithin{proposition}{section}
\numberwithin{definition}{section}
\numberwithin{lemma}{section}
\numberwithin{assumption}{section}
\numberwithin{remark}{section}

\def\1{\bm{1}}

\makeatletter
\let\save@mathaccent\mathaccent
\newcommand*\if@single[3]{%
    \setbox0\hbox{${\mathaccent"0362{#1}}^H$}%
    \setbox2\hbox{${\mathaccent"0362{\kern0pt#1}}^H$}%
    \ifdim\ht0=\ht2 #3\else #2\fi
}
\newcommand*\rel@kern[1]{\kern#1\dimexpr\macc@kerna}
\newcommand*\widebar[1]{\@ifnextchar^{{\wide@bar{#1}{0}}}{\wide@bar{#1}{1}}}
\newcommand*\wide@bar[2]{\if@single{#1}{\wide@bar@{#1}{#2}{1}}{\wide@bar@{#1}{#2}{2}}}
\newcommand*\wide@bar@[3]{%
    \begingroup
    \def\mathaccent##1##2{%
        \let\mathaccent\save@mathaccent
        \if#32 \let\macc@nucleus\first@char \fi
        \setbox\z@\hbox{$\macc@style{\macc@nucleus}_{}$}%
        \setbox\tw@\hbox{$\macc@style{\macc@nucleus}{}_{}$}%
        \dimen@\wd\tw@
        \advance\dimen@-\wd\z@
        \divide\dimen@ 3
        \@tempdima\wd\tw@
        \advance\@tempdima-\scriptspace
        \divide\@tempdima 10
        \advance\dimen@-\@tempdima
        \ifdim\dimen@>\z@ \dimen@0pt\fi
        \rel@kern{0.6}\kern-\dimen@
        \if#31
        \overline{\rel@kern{-0.6}\kern\dimen@\macc@nucleus\rel@kern{0.4}\kern\dimen@}%
        \advance\dimen@0.4\dimexpr\macc@kerna
        \let\final@kern#2%
        \ifdim\dimen@<\z@ \let\final@kern1\fi
        \if\final@kern1 \kern-\dimen@\fi
        \else
        \overline{\rel@kern{-0.6}\kern\dimen@#1}%
        \fi
    }%
    \macc@depth\@ne
    \let\math@bgroup\@empty \let\math@egroup\macc@set@skewchar
    \mathsurround\z@ \frozen@everymath{\mathgroup\macc@group\relax}%
    \macc@set@skewchar\relax
    \let\mathaccentV\macc@nested@a
    \if#31
    \macc@nested@a\relax111{#1}%
    \else
    \def\gobble@till@marker##1\endmarker{}%
    \futurelet\first@char\gobble@till@marker#1\endmarker
    \ifcat\noexpand\first@char A\else
    \def\first@char{}%
    \fi
    \macc@nested@a\relax111{\first@char}%
    \fi
    \endgroup
    }
\makeatother
\let\bar\widebar

\DeclareMathAlphabet{\mathsfit}{\encodingdefault}{\sfdefault}{m}{sl}
\SetMathAlphabet{\mathsfit}{bold}{\encodingdefault}{\sfdefault}{bx}{n}

\let\hat\widehat

\makeatletter

\def\foci@authors{}
\def\foci@affil{}
\def\foci@email{}
\def\foci@project{}
\def\foci@date{\today}
\def\foci@authorsep{}

\newcommand{\fociauthor}[2]{%
  \g@addto@macro\foci@authors{\foci@authorsep\mbox{#1\textsuperscript{#2}}\gdef\foci@authorsep{,\ }}}
\newcommand{\fociaffil}[1]{\gdef\foci@affil{#1}}
\newcommand{\fociemail}[1]{\gdef\foci@email{#1}}
\newcommand{\fociproject}[1]{\gdef\foci@project{#1}}
\newcommand{\focidate}[1]{\gdef\foci@date{#1}}
\newcommand{\fociauthorlist}{\gdef\foci@authorsep{}\foci@authors}

\def\foci@blfootnote{\gdef\@thefnmark{}\@footnotetext}
\newcommand{\focifootnotes}{%
  \foci@blfootnote{\textsuperscript{*}Equal contribution.\quad
    \textsuperscript{\Letter}Corresponding author: \texttt{\foci@email}}}

\newlength{\foci@mw}
\newcommand{\focidrawmark}[3]{%
  \setlength{\foci@mw}{#2}%
  \begin{scope}[shift={#3}, x={\foci@mw/564.03}, y={-\foci@mw/564.03}]
    \foreach \cx/\cy in {282.02/66.03, 282.02/222.08, 215.77/348.8, 348.73/348.8,
                         414.88/474.93, 149.16/474.93, 67.06/222.08, 496.99/222.08}
      \fill[#1] (\cx,\cy) circle[radius=42.04];
  \end{scope}}
\newcommand{\focimark}[2][black]{%
  \setlength{\foci@mw}{#2}%
  \begin{tikzpicture}[x={\foci@mw/564.03}, y={-\foci@mw/564.03}]
    \useasboundingbox (0,0) rectangle (564.03,540.96);
    \focidrawmark{#1}{#2}{(0,0)}
  \end{tikzpicture}}

\fancypagestyle{focititlepage}{\fancyhf{}\renewcommand{\headrulewidth}{0pt}}

\newcommand{\foci@headerline}{%
  \noindent
  \begin{minipage}[c]{0.5\linewidth}
    \includegraphics[height=0.58cm]{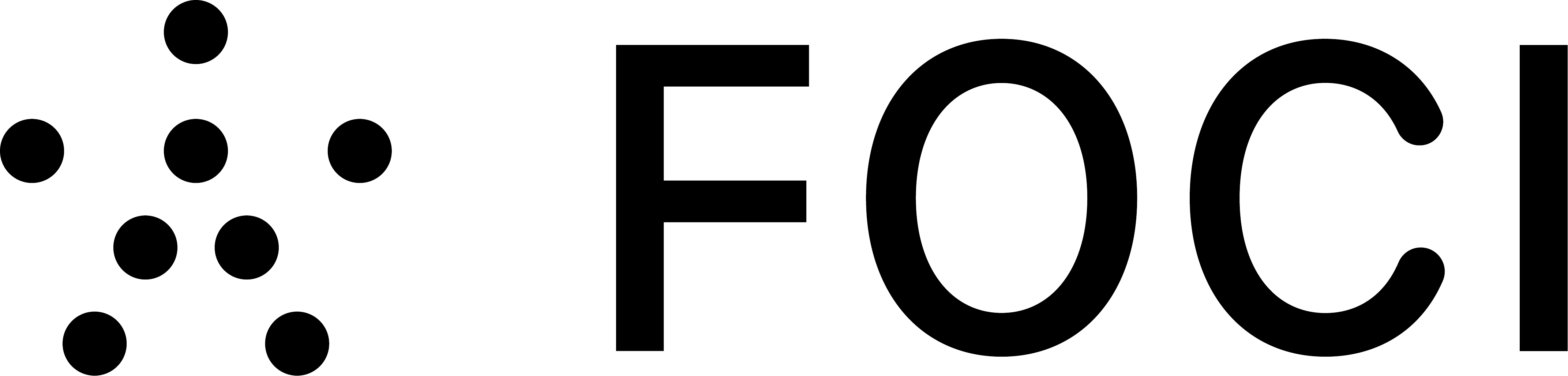}
  \end{minipage}%
  \begin{minipage}[c]{0.5\linewidth}
    \raggedleft\small\itshape\color{black!55}\foci@date
  \end{minipage}\par}

\newcommand{\foci@contactline}{%
  {\footnotesize\color{black!65}%
   \textbf{Email:}~\texttt{\foci@email}\qquad
   \textbf{Project:}~\url{\foci@project}\par}}

\newcommand{\foci@hairline}{\noindent\textcolor{black}{\rule{\linewidth}{0.5pt}}\par}

\newcommand{\focicoverCard}{%
  \thispagestyle{focititlepage}%
  \foci@headerline
  \vskip -0.5em
  \foci@hairline
  \vskip 0.9em
  \begin{tcolorbox}[enhanced, frame hidden, boxrule=0pt, arc=12pt,
      colback=black!4, width=\linewidth,
      left=18pt, right=18pt, top=18pt, bottom=16pt,
      grow to left by=2pt, grow to right by=2pt]
    \noindent
    \begin{minipage}[t]{\linewidth}
      {\setstretch{1.15}\LARGE\bfseries\raggedright Training and Scaling Compute-Optimal Physiological Waveform Foundation Models
\unskip\par}
      \vskip 0.55em
      {\normalsize\fociauthorlist\par}
      \vskip 0.2em
      {\small\color{black!70}\foci@affil\par}
    \end{minipage}\par
    \vskip 0.9em
    {\small\setstretch{1.08}We investigate the scaling laws and compute-optimal training of physiological
waveform foundation models (FMs). We train \aether{}, a family of over one hundred
FMs ranging from 20M to 2.1B parameters, on up to 36.3M hours of physiological waveforms. We construct eight clinical prediction tasks from
MIMIC-III and evaluate the FMs through linear probing. The 720M FM outperforms
all existing baseline FMs across all eight tasks. A scaling law of model size,
pretraining hours, and labeled patients predicts downstream ranking error, \textit{i.e.} $1-\auroc$, effectively with
$0.5\%$ prediction MAE at held-out resource scales and $0.9\%$ MAE when
extrapolating to 2.1B parameters. We present three findings:
(1) Compute-optimal training scales both FM size and pretraining hours.
Under the fitted law, a $10.0\times$ increase in compute FLOPs scales model size by
$1.2\times$ and pretraining hours by $8.2\times$.
(2) Larger FMs use waveform data more efficiently, and greater pretraining
exposure increases the benefit of model scaling. Starting from 25M parameters
and 4.8M pretraining hours, doubling FM size reduces the predicted hours needed
for the same performance by $51.8\%$.
(3) Pretraining and clinical supervision reinforce each other: more labeled
patients increase the return to pretraining, while larger FMs and longer
pretraining reduce labeling requirements. For the example of the 720M FM, extending
pretraining from 120K to 36.3M hours reduces the predicted patient requirement
by $61\%$ at a target ranking error.
These findings provide a quantitative training recipe and a promising and
durable scaling path for physiological waveform modeling and downstream clinical prediction.
\unskip\par}
    \vskip 0.9em
    \foci@contactline
  \end{tcolorbox}
  \vskip 0.4em}

\newcommand{\focicover}{\focicoverCard}

\makeatother

\usepackage{needspace}
\usepackage{tgcursor}
\graphicspath{{figures/}}
\newcommand{\aether}{\texttt{Aether}}
\newcommand{\auroc}{\mathrm{AUROC}}
\newcommand{\risk}{\mathcal{R}}
\definecolor{placeholder}{HTML}{875B35}
\newcommand{\TBD}[1]{\textcolor{placeholder}{\ensuremath{\mathrm{XXX}}}}

\hypersetup{
  pdftitle={Training and Scaling Compute-Optimal Physiological Waveform Foundation Models},
  pdfauthor={Pingzhi Li, Jie Peng, Shuqing Luo, Zachary Plotkin, Tianlong Chen},
  pdfsubject={Physiological waveform foundation models and scaling laws},
  colorlinks=true, linkcolor=teal, citecolor=teal, urlcolor=teal
}
\icmlpreprint{}
\titlespacing*{\section}{0pt}{2.2ex plus .5ex minus .2ex}{1ex plus .2ex}
\titlespacing*{\subsection}{0pt}{1.8ex plus .4ex minus .2ex}{.8ex plus .2ex}
\titlespacing*{\subsubsection}{0pt}{1.5ex plus .3ex minus .2ex}{.6ex plus .2ex}
\AtBeginEnvironment{thebibliography}{\raggedright}

\makeatletter
\patchcmd{\BR@backref}{\newblock}{\unskip\nobreak\hspace{0.5em}}{}{}
\renewcommand{\foci@contactline}{}
\renewcommand{\focifootnotes}{%
  \foci@blfootnote{\textsuperscript{\ensuremath{\dagger}}Corresponding authors:
  Zachary Plotkin and Tianlong Chen (\href{mailto:tianlong@cs.unc.edu}{tianlong@cs.unc.edu}).}}
\makeatother

\begin{document}
\fociauthor{Pingzhi Li}{2}
\fociauthor{Jie Peng}{2}
\fociauthor{Shuqing Luo}{2}
\fociauthor{Zachary Plotkin}{1,\ensuremath{\dagger}}
\fociauthor{Tianlong Chen}{1,2,\ensuremath{\dagger}}

\fociaffil{\textsuperscript{1}Foci Labs\quad
\textsuperscript{2}University of North Carolina at Chapel Hill}

\pagestyle{fancy}
\fancyhf{}
\renewcommand{\headrulewidth}{0.4pt}
\chead{\small\scshape Aether: Compute-Optimal Physiological Waveform FMs
\unskip}
\cfoot{\small\thepage}

\focicover
\focifootnotes
\section{Introduction}
\label{sec:introduction}

\begin{figure}[!ht]
\centering
\includegraphics[width=\linewidth]{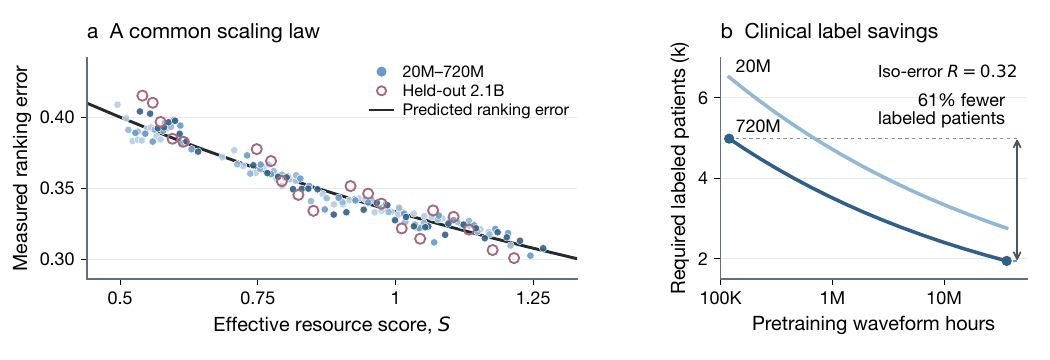}
\caption{Scaling clinical prediction and label efficiency.
(a) Measured eight-task mean ranking error against
$S=P^g/(AN^{-a}+BD^{-b})$.
(b) Predicted labeled patient requirements at $\risk=0.32$.
Extending 720M pretraining from 120K to 36.3M hours reduces required
patients by $61\%$.}
\label{fig:intro-scaling-teaser}
\end{figure}

Scaling laws for large language models describe how language modeling loss
follows power-law relationships with model size and training
data~\citep{kaplan2020scaling}, and compute-optimal training scales model size and
token count in approximately equal proportion as compute
grows~\citep{hoffmann2022chinchilla}. The practical value is that a large training
run can be configured in advance, allocating its compute before committing to
the most expensive experiments. Physiological waveforms bring this
resource-allocation question to continuous measurements and clinical supervision.
Existing work offers two complementary starting points. Foundation models
demonstrate transferable representations for electrocardiography,
photoplethysmography, electroencephalography, sleep polysomnography, and
multimodal physiological signals~\citep{mckeen2024ecgfm,pillai2025papagei,wang2025cbramod,thapa2026sleepfm,guo2025qualityfm}.
Scaling studies characterize resource-performance relationships in structured
clinical event records~\citep{zhang2025ehrscaling},
ECG~\citep{li2026ecgscaling,almasud2026ecgpretraining}, and wearable and
intracranial recordings~\citep{narayanswamy2024wearablescaling,han2026diver}.
These advances motivate a scaling law of FM size, waveform exposure, and labeled patients.

\looseness=-1
We ask: \emph{Given a pretraining compute budget and a labeled-patient budget,
how should model size and waveform exposure scale to achieve the best downstream clinical performance?} To study this question, we train \aether{}, a family of over one
hundred foundation models~(FMs) ranging from 20M to 2.1B parameters, on up to
36.3M cumulative hours of physiological waveforms. To efficiently model long waveform contexts, we use a
bidirectional hybrid backbone based on Gated DeltaNet
(GDN)~\citep{yang2025gateddeltanet}, arranging GDN and full-attention blocks in a
repeating $3{:}1$ pattern as in \citet{qwen2026qwen35}.
We construct eight clinical prediction tasks from
MIMIC-III waveforms and clinical
records~\citep{moody2020mimicwaveform,johnson2016mimic}. Each pretrained FM
is evaluated with a linear probe on frozen representations, following
established protocols in visual representation
learning~\citep{chen2020simclr,he2020moco,he2022mae} and ECG representation
learning~\citep{kiyasseh2021clocs,diamant2022pclr}. Overall, we evaluate
over one thousand combinations of model size, pretraining hours, labeled-patient
counts, and clinical task, studying how
resources invested both before and after pretraining translate into downstream physiological performance.
Our \aether{}-720M outperforms the existing baseline FMs across all eight tasks.

We utilize this resource study to derive a quantitative training recipe.
A scaling law relates the eight-task mean ranking error to FM size $N$,
pretraining waveform hours $D$, and labeled patients per task $P$
(Figure~\ref{fig:intro-scaling-teaser}a).
It predicts held-out resource configurations with $0.5\%$ MAE and
generalizes to larger FMs and longer pretraining runs.
We use this law to derive compute-optimal model--data allocations
and quantify the waveform and labeling requirements for a target
clinical performance.

We present three findings: (1) Compute-optimal training scales FM size and
pretraining hours together, with hours growing faster than model size.
Under the fitted law, a $10.0\times$ increase in compute scales FM size by
$1.2\times$ and pretraining hours by $8.2\times$.
(2) Larger FMs are more data-efficient, and longer pretraining amplifies
model-scaling gains. At fixed $P$ and ranking error, doubling a 25M FM
from a 4.8M-hour reference reduces predicted pretraining hours by $51.8\%$.

(3) More labeled patients amplify pretraining gains, while scaling
pretraining reduces labeling needs. For example, the law predicts $27.2\%$
fewer labeled patients with larger FMs and $61\%$ fewer with longer
pretraining at fixed target performance
(Figure~\ref{fig:intro-scaling-teaser}b; Section~\ref{sec:prediction}).
In practice, clinical labels add expert curation to the access requirements
of raw waveforms, making both harder to expand than
compute~\citep{krishnan2022selfsupervised,rieke2020federated}.
This suggests a promising and durable scaling path: scale pretraining on
available waveforms to improve clinical prediction with fewer labels.

Our contributions and findings are summarized as follows:
(1) We establish a predictive scaling law for clinical performance and derive
compute-optimal allocations for scaling FM size and pretraining hours with compute.
(2) We quantify the data-efficiency benefits of model scaling: larger FMs
reach the same predicted performance with fewer waveform hours, while greater
pretraining exposure increases the benefit of scaling model size.
(3) We characterize how pretraining and clinical supervision reinforce each
other: more labeled patients increase the return to pretraining, while larger
FMs and longer pretraining reduce the predicted labeling requirements for
a target clinical performance.

\section{Related work}

\paragraph{Scaling and downstream transfer.}
Empirical studies characterize predictable relationships between dataset size,
model capacity, and generalization error across learning
domains~\citep{hestness2017predictable,rosenfeld2020constructive}.
Language-model scaling laws connect model size, data, and
compute~\citep{kaplan2020scaling}, with compute-optimal training jointly
allocating parameters and training tokens~\citep{hoffmann2022chinchilla}.
Related analyses extend to vision transformers~\citep{zhai2022scaling}
and neural machine translation~\citep{ghorbani2022nmt}.
Transfer scaling expresses pretraining benefits through downstream data
requirements~\citep{hernandez2021transfer}, while data-constrained scaling
distinguishes new tokens from repeated exposure~\citep{muennighoff2023dataconstrained}.
Studies of extrapolation and changes in scaling regimes further examine which
functional forms predict performance beyond the fitted
range~\citep{alabdulmohsin2022revisiting,caballero2023broken}.
These works motivate our scaling model and its evaluation at unseen
model scales and compute allocations for clinical prediction.

\paragraph{Scaling in physiological and clinical data.}
\citet{li2026ecgscaling} jointly examine ECG architecture, pretraining paradigm,
and model--data scaling, while \citet{almasud2026ecgpretraining} compare
self-supervised objectives as ECG pretraining data grows.
Wearable scaling studies distinguish subject count from recording
duration~\citep{narayanswamy2024wearablescaling}; DIVER-1 studies model size,
unique data, subject count, and training duration for
iEEG~\citep{han2026diver}. Complementary analyses consider modality diversity
and model capacity in nocturnal biosignals~\citep{yuan2026sleep2vec}, data per
subject in brain-to-image decoding~\citep{banville2025brainscaling}, and
compute-optimal scaling in EHR sequences~\citep{zhang2025ehrscaling}.
Cohort-composition experiments also show that ECG transfer depends on the
distribution of pretraining patients~\citep{khattak2026distribution}.
Building on these precedents, we study a joint model--exposure--supervision
law for clinical waveform prediction, with labeled patients as an explicit
resource coordinate. We compare patient- and window-based label acquisition
under a common model ladder and evaluation protocol to test whether that
coordinate transfers across acquisition designs while keeping task definitions,
validation cohorts, and selection criteria fixed.

\paragraph{Physiological waveform foundation models.}
Early ECG pretraining methods, including CLOCS~\citep{kiyasseh2021clocs} and
PCLR~\citep{diamant2022pclr}, show that unlabeled cardiac recordings can
support label-efficient downstream learning.
ECG-FM combines contrastive and generative pretraining for electrocardiogram
analysis~\citep{mckeen2024ecgfm}. PaPaGei uses waveform morphology to learn
transferable photoplethysmography representations~\citep{pillai2025papagei}.
For EEG, BENDR learns transferable features through contrastive
pretraining~\citep{kostas2021bendr}. LaBraM learns by predicting masked neural
tokens~\citep{jiang2024labram}, while CBraMod uses masked reconstruction with
separate spatial and temporal attention~\citep{wang2025cbramod}.
BIOT supports learning across biosignal datasets with heterogeneous channels,
sample lengths, and missing values~\citep{yang2023biot}.
SleepFM learns multimodal polysomnography representations for disease
prediction~\citep{thapa2026sleepfm}, and QualityFM addresses variable ECG and
PPG signal quality through self-distillation~\citep{guo2025qualityfm}.
These studies establish the utility of pretrained waveform representations
across modalities and clinical tasks. Our focus is to quantify how resources
allocated to pretraining and downstream supervision translate into clinical
performance under a common linear-probe protocol across all model sizes,
pretraining exposures, and labeled-patient budgets.

\section{Experimental design}
\label{sec:design}

We study scaling by varying FM size, pretraining hours, and labeled-patient
coverage within a common training and evaluation framework.
Section~\ref{sec:architecture} presents the \aether{} architecture and
scaling ladder. Section~\ref{sec:pretraining} describes the waveform data
and pretraining recipe. Section~\ref{sec:linear-probing} introduces the
eight clinical tasks, downstream linear-probing protocol, and evaluation metrics.

\begin{figure}[t]
\centering
\includegraphics[width=1.0\linewidth,pagebox=cropbox]{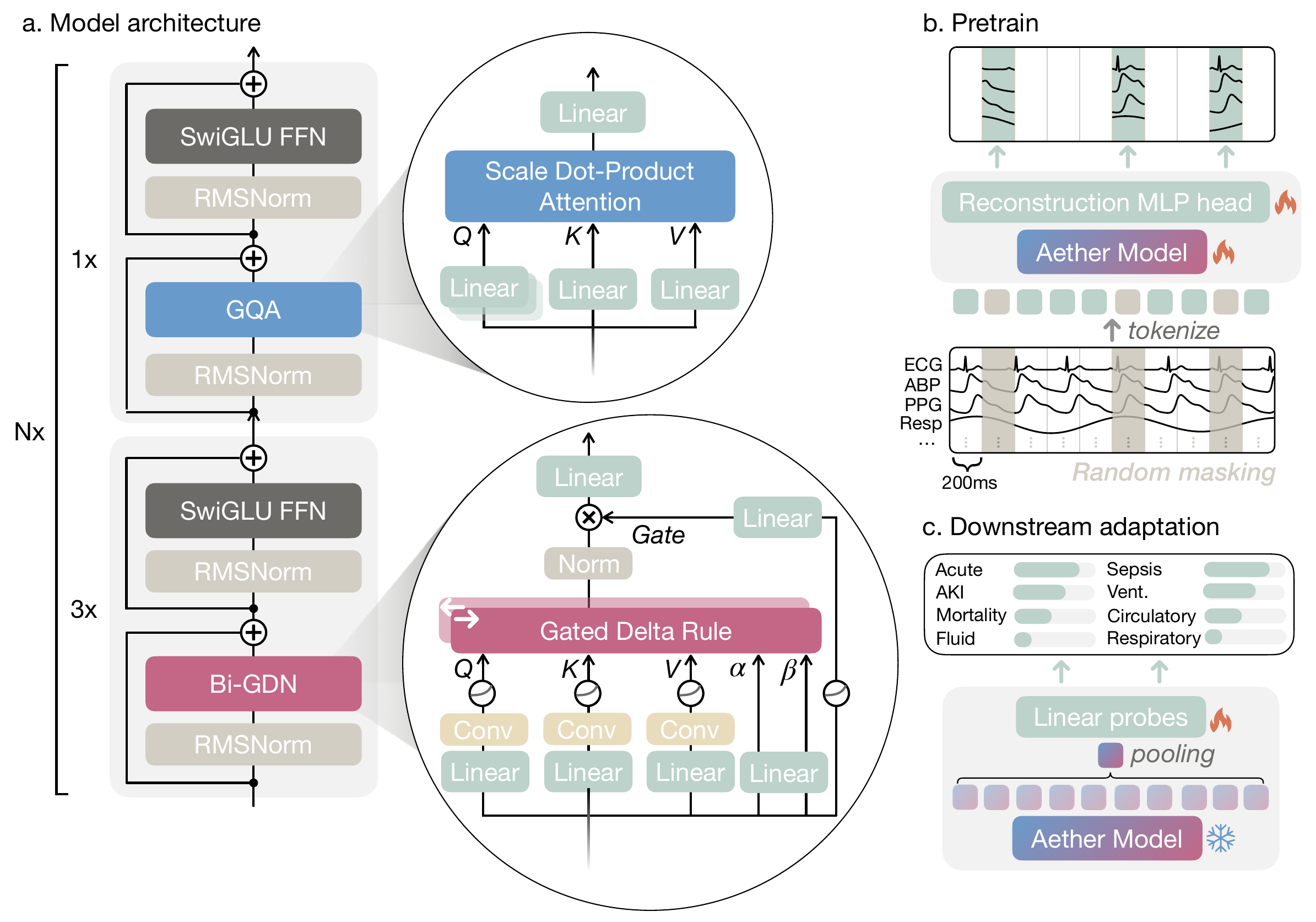}
\caption{Overview of \aether{} architecture, pretraining, and downstream
adaptation. (a) Bidirectional Gated DeltaNet (Bi-GDN) and grouped-query
attention (GQA) blocks form a repeating $3{:}1$ hybrid pattern.
(b) Multichannel waveforms are divided into 200\,ms patches, and the FM
learns to reconstruct randomly masked patches.
(c) The pretrained FM is frozen, and its mean-pooled representations
support linear probing for downstream clinical prediction.}
\label{fig:model-arch}
\end{figure}

\subsection{Model architecture}
\label{sec:architecture}

\newsavebox{\scalingladderbox}
\sbox{\scalingladderbox}{\footnotesize\setlength{\tabcolsep}{3pt}%
\begin{tabular}{@{}lrrrrr@{}}
\toprule
\midrule
Size & Width & Depth & FFN & Q/KV & GDN \\
\midrule
20M & 640 & 5 & 2,048 & 5/1 & 1 \\
60M & 896 & 7 & 2,816 & 7/1 & 2 \\
180M & 1,280 & 10 & 4,096 & 10/2 & 3 \\
720M & 1,920 & 15 & 6,144 & 15/3 & 4 \\
2.1B & 2,816 & 22 & 8,960 & 22/2 & 6 \\
\midrule
\bottomrule
\end{tabular}
}
\begin{wraptable}{r}{\wd\scalingladderbox}
\centering\footnotesize
\setlength{\tabcolsep}{3pt}
\setlength{\abovecaptionskip}{0pt}
\setlength{\belowcaptionskip}{4pt}
\caption{\aether{} scaling ladder. FFN: Dimension of FFN; Q/KV: full-attention head counts; GDN: recurrent heads.}
\label{tab:scaling-ladder}
\usebox{\scalingladderbox}
\end{wraptable}
Figure~\ref{fig:model-arch} summarizes the training and adaptation of
\aether{}, which tokenizes one hour of physiological waveforms into a sequence of
continuous representations. Inputs are sampled at 125\,Hz and 62 channels. We normalize each channel
within its input window and jointly project patches of 25 samples across
channels, giving 18,000 temporal tokens per hour.
To model this long context efficiently, the FM interleaves bidirectional
Gated DeltaNet blocks~\citep{yang2025gateddeltanet} and full-attention blocks
in a repeating $3{:}1$ pattern~\citep{qwen2026qwen35}. Full-attention
blocks use no positional encoding (NoPE)~\citep{kazemnejad2023positional}. Each block uses RMSNorm~\citep{zhang2019rmsnorm}
and a SwiGLU feed-forward layer~\citep{shazeer2020glu}.

The scaling ladder derives hidden width, FFN width, and head counts from
depth, as shown in Table~\ref{tab:scaling-ladder}, with a same 3:1 hybrid block pattern. Full-attention and GDN heads have dimension
128, and the GDN convolution width is 4. We use the four smaller FMs to establish the scaling laws, while
\aether{}-2.1B provides the held-out model-size evaluation. Scaling fits use the exact FM parameter count $N$, excluding the reconstruction MLP head.

\subsection{Pretraining}
\label{sec:pretraining}

\paragraph{Data.} The pretraining corpus comprises 398,204 hours of inputs from MIMIC-III
waveform recordings~\citep{moody2020mimicwaveform,johnson2016mimic}. We pretrain by random masked
reconstruction~\citep{he2022mae}, where each temporal patch is independently
masked with probability 0.30, replacing its projected token with a learned
mask embedding. A reconstruction head predicts the normalized waveform,
and the loss averages squared error over observed values in masked patches:
\begin{equation}
\mathcal{L}_{\mathrm{PT}}
=\frac{\sum_j m_j o_j(\hat x_j-x_j)^2}{\sum_j m_j o_j},
\label{eq:pretraining-loss}
\end{equation}
where $j$ indexes scalar samples across the minibatch, $o_j$ indicates an
observed value, and $m_j$ indicates that its temporal patch is masked.
We use AdamW~\citep{loshchilov2019adamw} with a global batch of 512,
learning rate $10^{-4}$, $(\beta_1,\beta_2)=(0.9,0.95)$, weight decay 0.05,
and gradient-norm clipping at 1.0. We follow the warmup-stable-decay
(WSD) schedule~\citep{hu2024minicpm}, with 100 warmup steps, a stable
learning rate of $10^{-4}$, and 100 cosine-decay steps to $10^{-5}$.
Every pretraining run has its own terminal decay to ensure convergence.

\paragraph{Training.} We measure pretraining exposure $D$ in waveform-hours and evaluate
119K, 239K, 398K, 796K, 1.59M, and 2.39M hours.
To extend pretraining hours beyond the available data, we repeat the
pretraining waveforms when needed, with $E$ denoting the pretraining repeat
ratio. Compute budgets include forward and
backward FLOPs accumulated over pretraining updates, following \citet{hoffmann2022chinchilla}.
We also built a JAX-based~\citep{jax2018github} distributed training infra, named FoXus, for our internal heterogeneous clusters spanning AWS Trainium2, Google TPU v5p, and
NVIDIA B200. We built FoXus from scratch to give us full end-to-end control over the training stack, including compiler, kernels, parallelism and scheduling~\footnote{We achieve MFU at 14\% for Trainium2, 40\% for v5p, and 34\% for B200.}.

\subsection{Downstream linear probing}
\label{sec:linear-probing}

\newsavebox{\cohorttablebox}
\sbox{\cohorttablebox}{\footnotesize\setlength{\tabcolsep}{2.2pt}%
% Accepted cohort counts; hours sum nominal one-hour input durations.
\begin{tabular}{@{}lrrrrrr@{}}
\toprule
\midrule
& \multicolumn{2}{c}{Train} & \multicolumn{2}{c}{Validation} & \multicolumn{2}{c}{Test} \\
\cmidrule(lr){2-3}\cmidrule(lr){4-5}\cmidrule(lr){6-7}
Task & Hours & Patients & Hours & Patients & Hours & Patients \\
\midrule
Acute & 112,727 & 5,088 & 13,852 & 623 & 14,683 & 637 \\
Circulatory & 316,881 & 4,785 & 37,109 & 595 & 463,197 & 877 \\
Mortality & 593,154 & 7,112 & 72,226 & 885 & 77,809 & 887 \\
Respiratory & 122,597 & 4,334 & 14,937 & 545 & 16,356 & 542 \\
Sepsis & 152,486 & 3,892 & 22,089 & 480 & 214,884 & 726 \\
Fluid response & 4,742 & 1,491 & 552 & 176 & 692 & 205 \\
KDIGO 2/3 AKI & 124,623 & 3,800 & 15,315 & 461 & 14,542 & 481 \\
Vent. readiness & 9,276 & 2,069 & 1,112 & 240 & 1,215 & 245 \\
\midrule
\bottomrule
\end{tabular}
\unskip}
\begin{wraptable}{r}{\wd\cohorttablebox}
\centering\footnotesize
\setlength{\abovecaptionskip}{0pt}
\setlength{\belowcaptionskip}{4pt}
\caption{Overview of downstream task datasets.}
\label{tab:cohorts}
\usebox{\cohorttablebox}
\end{wraptable}
\paragraph{Data.} We construct eight held-out clinical prediction tasks from MIMIC-III waveforms and
clinical records~\citep{moody2020mimicwaveform,johnson2016mimic} to evaluate the FMs' performance: acute deterioration, circulatory deterioration, dynamic mortality, respiratory escalation, sepsis onset, fluid response, KDIGO stage 2/3 acute kidney injury, and ventilator readiness within 24 hours.
Acute deterioration has three binary targets, fluid response and acute
kidney injury have two each, and the other tasks have one.
Patients in all downstream tasks are held out from the pretraining cohort.
Table~\ref{tab:cohorts} summarizes the training, validation and test
cohorts.
Details of task definitions and construction are in Appendix~\ref{app:clinical-downstream-datasets}.

\paragraph{Training.} For linear probing, we sample \{10\%, 30\%, 60\%, 100\%\} of the
full training patients, including
all inputs from each selected patient. We record
the labeled-patient count $P$ as the patient-scaling coordinate.
Together, the FM sizes, pretraining hours, patient counts, and tasks
yield about one thousand combinations. A companion analysis samples
the same fractions of waveform windows.

Following standard linear probing
protocols~\citep{chen2020simclr,he2020moco,he2022mae}, we freeze the FM,
mean-pool its representations, and fit an $L_2$-regularized logistic probe for each target. Features are standardized using the selected training subset. Five candidates use fixed regularization strengths
$\lambda\in\{0.001,0.01,0.1,1,10\}$.
We select one complete multi-target probe per condition by validation
patient-balanced macro AUROC and use it for every reported metric.

\label{sec:metrics}
\paragraph{Metrics.} Our primary outcome is ranking error,
$\risk=1-\bar A$~\citep{agarwal2014ranking}, where $\bar A$ is
patient-balanced AUROC averaged across the task's targets. Each patient's
observed inputs receive equal total weight per
target~\citep{williamson2003marginal,kahan2023estimands}.
Changes in ranking error are absolute shifts on the percentage scale.
We estimate uncertainty with 10,000 paired patient-bootstrap draws,
retaining all inputs from each sampled
patient~\citep{obuchowski1997clusteredroc,li2008clusteredroc}.

\section{Scaling Laws and Evaluation}
\label{sec:results}

\newsavebox{\scalingnotationbox}
\sbox{\scalingnotationbox}{\footnotesize\setlength{\tabcolsep}{4pt}%
\begin{tabular}{@{}ll@{}}
\toprule
\midrule
Symbol & Description \\
\midrule
$\bar A$ & Patient-level macro AUROC \\
$\risk$ & Ranking error, $1-\bar A$ \\
$\widehat{\risk}$ & Predicted ranking error \\
$N$ & FM parameter count \\
$D$ & Pretraining waveform hours \\
$P$ & Labeled patients per task \\
$C$ & Pretraining compute (FLOPs) \\
$q$ & Waveform tokens per hour \\
\midrule
\bottomrule
\end{tabular}
}
\begin{wraptable}{r}{\wd\scalingnotationbox}
\centering\footnotesize
\setlength{\abovecaptionskip}{0pt}
\setlength{\belowcaptionskip}{4pt}
\caption{Summary of notations.}
\label{tab:scaling-notation}
\usebox{\scalingnotationbox}
\end{wraptable}
\looseness=-1
Table~\ref{tab:scaling-notation} summarizes the scaling-law notation.
This section connects the clinical performance of \aether{} to model
capacity, pretraining exposure, and downstream supervision.
Section~\ref{sec:baseline-comparison} establishes the effectiveness of our
pretraining recipe through comparisons with random initialization and
existing FMs. Section~\ref{sec:scaling-law} fits a scaling law and evaluates
its predictions at held-out resource levels.
Section~\ref{sec:scale-extrapolation} evaluates predictions at 2.1B and
longer pretraining durations.

\subsection{Comparison with Existing FMs}
\label{sec:baseline-comparison}

We first evaluate whether our pretraining recipe yields useful representations
for clinical prediction and how \aether{} compares with existing FMs.
This comparison establishes the clinical benefit of our training recipe
before examining its scaling behavior.

\paragraph{Baseline waveform FMs.}
We compare with thirteen pretrained models from six families:
TimesFM 2/2.5/3~\citep{das2024timesfm} and
Chronos-Bolt~\citep{ansari2024chronos}, developed for time-series forecasting;
MOMENT~\citep{goswami2024moment}, which learns general-purpose time-series
representations through masked reconstruction; SOTER~\citep{chen2026soter},
a generative FM for wearable physiological signals; and
NormWear 1/2~\citep{luo2026normwear,luo2026normwear2}, which learn wearable-signal
representations and physiological dynamics.
We also include randomly initialized \aether{} FMs at each size to measure
the contribution of our pretraining recipe. All FMs are frozen and evaluated with
linear probes using the full training cohorts and the same patient-balanced
validation metric. 

\begin{figure}[t]
\centering
\includegraphics[width=1.0\linewidth]{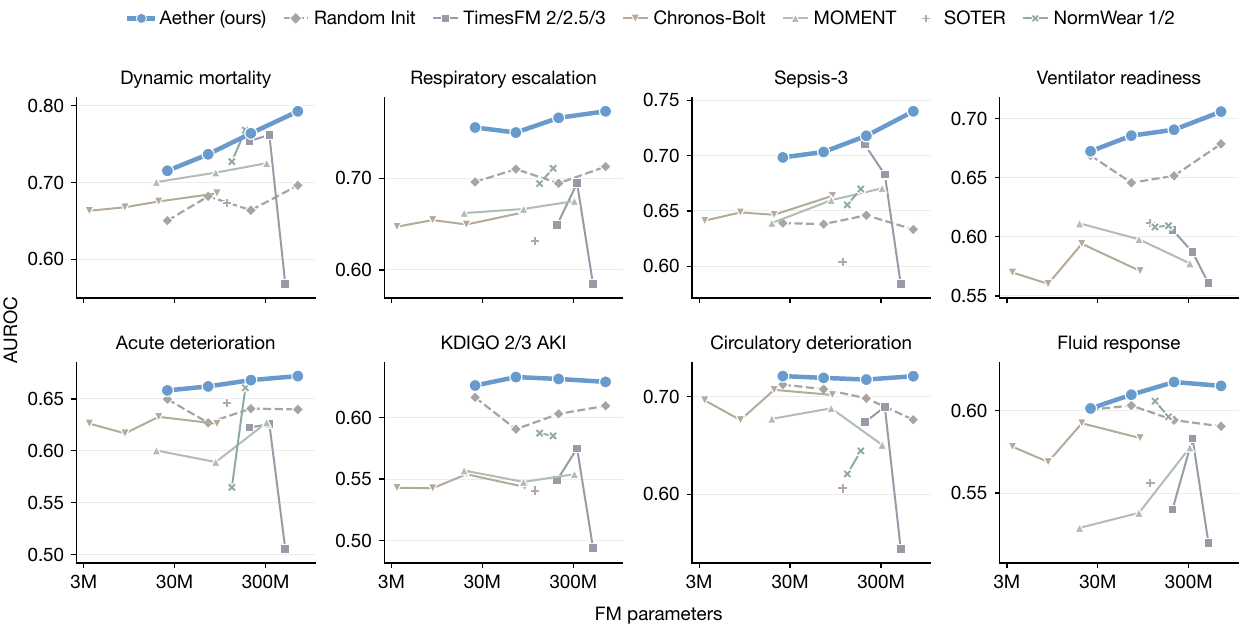}
\caption{Comparison with existing waveform FMs across eight clinical prediction tasks.
Each panel shows macro AUROC against FM size using 100\% of labeled patients.
To ensure a fair comparison, we report each model's best linear-probe AUROC
across the same regularization sweep,
$\lambda\in\{0.001,0.01,0.1,1,10\}$. Higher AUROC corresponds to lower
ranking error.}
\label{fig:baseline-comparison}
\end{figure}

\paragraph{Our pretraining improves clinical prediction.}
As shown in Figure~\ref{fig:baseline-comparison},
pretrained \aether{} outperforms its randomly initialized counterpart
across all eight tasks and four model sizes. At 720M, our pretraining
recipe improves eight-task mean AUROC by $6.4\%$, including gains of
$9.6\%$ for mortality and $10.7\%$ for sepsis. These gains demonstrate
the clinical value of our pretraining recipe.

\paragraph{\aether{} outperforms the evaluated baselines.}
The 720M FM achieves higher AUROC than every public baseline on each of
the eight tasks in Figure~\ref{fig:baseline-comparison}. Relative to the
strongest baseline within each task, ranking error decreases by
$0.9\%$--$9.4\%$, averaging $3.6\%$ across tasks. The reductions reach
$4.1\%$ for AKI, $6.2\%$ for respiratory escalation, and $9.4\%$ for
ventilator readiness. These results establish the effectiveness of the
\aether{} training recipe across diverse clinical endpoints. We next
quantify how its performance changes with FM size, pretraining hours,
and labeled patients.

\subsection{Fitting Scaling Laws}
\label{sec:scaling-law}

\begin{figure}[t]
\centering
\includegraphics[width=\linewidth]{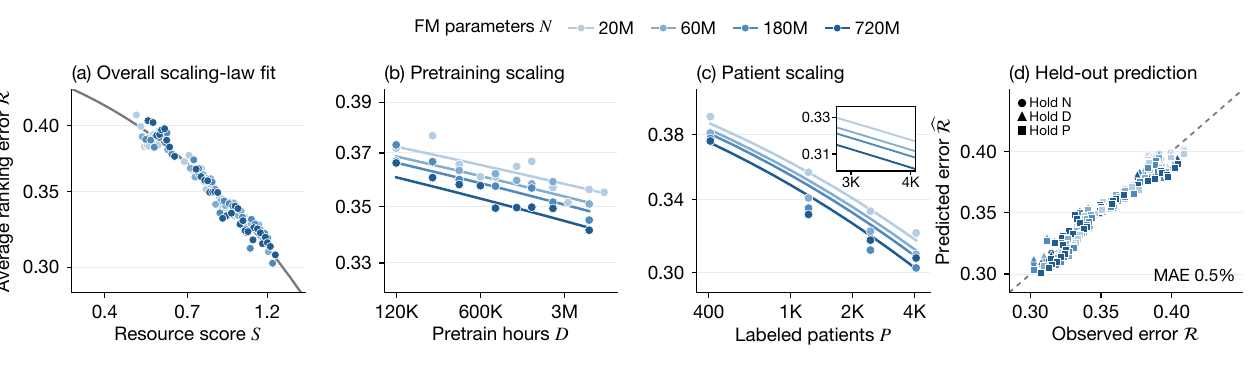}
\caption{Scaling of downstream average ranking error $\risk$ across eight tasks.
(a) Fitted response to the resource score $S=P^g/(AN^{-a}+BD^{-b})$.
(b) Pretraining scaling at $P=1{,}222$.
(c) Patient scaling at 4.8M pretraining hours.
(d) Predictions at held-out resource levels: we hold out all configurations
at one value of $N$, $D$, or $P$, fit on the rest, and predict the held-out
configurations. MAE averages absolute prediction errors equally across
the three resource axes.}
\label{fig:scaling-law-overview}
\end{figure}

We fit a scaling law for the eight-task mean ranking
error~\citep{kaplan2020scaling,hoffmann2022chinchilla}:
\begin{equation}
\widehat{\risk}=
\left[2+\frac{P^g}{AN^{-a}+BD^{-b}}\right]^{-1},
\label{eq:ndp-scaling-law}
\end{equation}
where $N$ is the FM parameter count, $D$ is pretraining waveform hours,
and $P$ is the number of labeled patients per task.
Nonlinear least squares over the resource grid gives
$A=6931.568$, $B=19.421$, $a=0.513$, $b=0.050$, and $g=0.296$.
Appendix~\ref{app:expected-forms} develops the expected scaling terms
and describes the fitting procedure.
Figure~\ref{fig:scaling-law-overview}(a) shows that configurations with
different $N$, $D$, and $P$ follow a common response curve when expressed
through the resource score $P^g/(AN^{-a}+BD^{-b})$.
Positive exponents give consistent improvements with each resource.
Across the fitting grid, ranking error responds most strongly to
proportional increases in labeled patients, identifying supervision as
the leading downstream scaling factor after pretraining.

Figure~\ref{fig:scaling-law-overview}(b,c) presents the effects of
pretraining hours and labeled patients respectively. The fitted ranking error curves consistently decrease as
either resource grows across all FM sizes, with larger FMs reaching
lower ranking error at the same $D$ and $P$.
The law also quantifies their greater data efficiency: at fixed $P$ and
target ranking error, $AN^{-a}+BD^{-b}$ remains constant.
Starting from 25M parameters and 4.8M pretraining hours, the fit predicts
that doubling FM size reduces the hours needed for the same error by
$51.8\%$.
These trends demonstrate consistent improvements from scaling FM size,
pretraining waveform hours, and labeled patients.

We evaluate predictions at unseen resource levels by leaving out all
configurations at one value of $N$, $D$, or $P$ in turn and fitting on
the remaining configurations. Figure~\ref{fig:scaling-law-overview}(d) shows close
agreement between predicted and observed ranking error, with a development
MAE of $0.5\%$. This reduces MAE by $50.1\%$ relative to a baseline that
uses the measured error of the nearest configuration available during
fitting, with distance measured in $(\log N,\log D,\log P)$ space.
These results demonstrate accurate prediction of ranking error at held-out
resource levels.

\subsection{Evaluation on Predicting Larger Scales}
\label{sec:scale-extrapolation}

\begin{figure}[t]
\centering
\includegraphics[width=\linewidth]{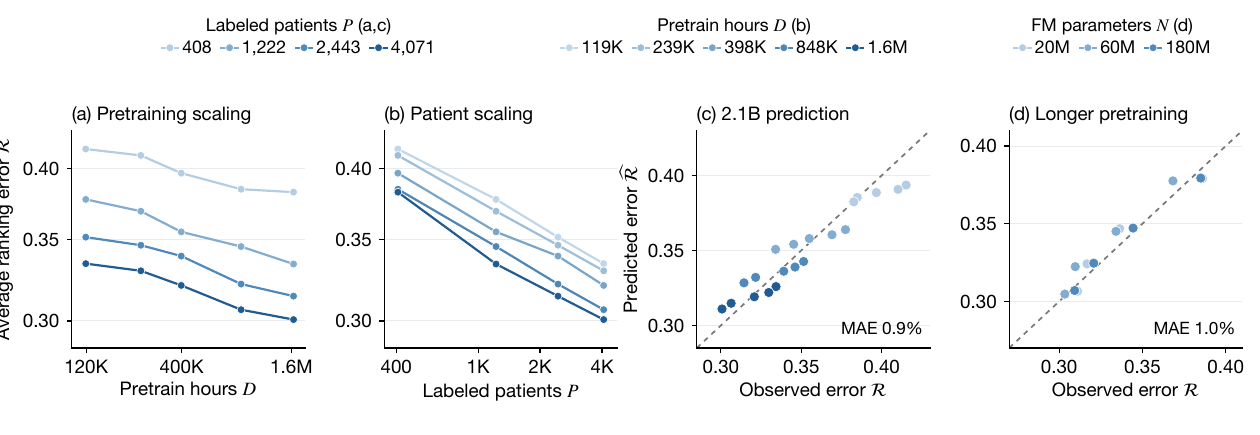}
\caption{\looseness=-1 Scaling and prediction at larger resources.
(a) Pretraining scaling at 2.1B for all four values of $P$.
(b) Patient scaling at 2.1B for all five values of $D$.
(c) Predictions for all 20 configurations at 2.1B.
(d) Predictions for 20M, 60M, and 180M after 36.3M, 17.0M, and 5.9M
pretraining waveform hours, respectively.
Ranking errors and absolute patient counts are averaged across the eight
tasks. Solid lines connect measurements in (a,b); dashed diagonals in
(c,d) mark $\widehat{\risk}=\risk$.}
\label{fig:larger-scale-evaluation}
\end{figure}

To further evaluate the generalizability of our scaling law, we apply
Equation~\ref{eq:ndp-scaling-law}, fitted on 20M--720M FMs,
to \aether{}-2.1B and to longer pretraining runs at 20M, 60M, and 180M.
Figure~\ref{fig:larger-scale-evaluation} summarizes all 20 configurations
at 2.1B and all 12 configurations at the longer training durations,
each evaluated across the same eight clinical tasks.

The gains from waveform pretraining and labeled-patient supervision extend
to the 2.1B scale. Increasing pretraining from 119K to 1.6M waveform hours
reduces average ranking error by $3.3\%$--$4.3\%$ across all four
labeled-patient counts (Figure~\ref{fig:larger-scale-evaluation}a).
Increasing $P$ from 408 to 4,071 reduces average ranking error by
$7.6\%$--$8.2\%$ across all five pretraining settings
(Figure~\ref{fig:larger-scale-evaluation}b).
These gains demonstrate consistent improvements from scaling our training
recipe at 2.1B.

The law predicts performance at 2.1B with an MAE of $0.9\%$ across all
20 configurations (Figure~\ref{fig:larger-scale-evaluation}c).
It also predicts the performance of 20M, 60M, and 180M FMs after 36.3M,
17.0M, and 5.9M pretraining waveform hours, respectively, with an overall
MAE of $1.0\%$ across the 12 configurations
(Figure~\ref{fig:larger-scale-evaluation}d).
On these longer runs, the law reduces MAE by $18.1\%$ relative to
nearest-configuration prediction.
These results demonstrate the scaling law's ability to generalize through
accurate extrapolation to larger FMs and longer pretraining durations.

\section{Compute-Optimal Training of Physiological FMs}
\label{sec:prediction}

\begin{figure}[!ht]
\centering
\includegraphics[width=\linewidth]{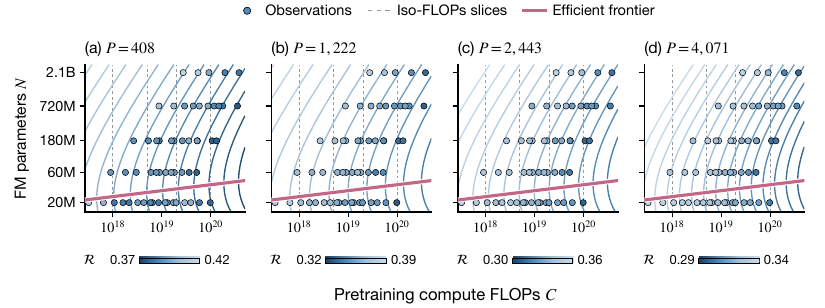}
\caption{Iso-error contours under a FLOPs constraint, from
Equation~\ref{eq:ndp-compute-law}.
(a--d) $P=408$, $1{,}222$, $2{,}443$, and $4{,}071$, respectively.
Dashed lines mark iso-FLOPs slices; red curves show the efficient frontier
over the observed FM-size range. Each panel shows all 57 configurations.
Ranking errors and patient counts are eight-task means.}
\label{fig:compute-efficient-frontier}
\end{figure}

\paragraph{Q1: How should model size and waveform data scale with compute?}
We express Equation~\ref{eq:ndp-scaling-law} under the training-compute
approximation $C\simeq6qND$~\citep{kaplan2020scaling,hoffmann2022chinchilla},
where $q=18{,}000$ waveform tokens per hour for \aether{}.
Substituting $D=C/(6qN)$ gives
\begin{equation}
\widehat{\risk}(C,N,P)=
\left[2+\frac{P^g}{AN^{-a}+B\left(C/(6qN)\right)^{-b}}\right]^{-1}.
\label{eq:ndp-compute-law}
\end{equation}
At fixed $P$, minimizing ranking error is equivalent to minimizing
$AN^{-a}+BD^{-b}$. The interior \mbox{compute-optimal allocation is}
\begin{align}
N^\star(C)&=\left(\frac{aA}{bB}\right)^{\!1/(a+b)}
\left(\frac{C}{6q}\right)^{\!b/(a+b)}\propto C^{0.088},
\label{eq:compute-size-optimum}\\
D^\star(C)&=\left(\frac{bB}{aA}\right)^{\!1/(a+b)}
\left(\frac{C}{6q}\right)^{\!a/(a+b)}\propto C^{0.912}.
\label{eq:compute-hours-scaling}
\end{align}
Figure~\ref{fig:compute-efficient-frontier} shows this allocation across
all four labeled-patient cohorts. For interior solutions, a $10\times$ increase
in compute scales FM size by $1.2\times$ and pretraining hours by
$8.2\times$. These predictions provide two ways to use additional compute.
\ding{182} Scale FM size and pretraining waveform hours together,
with waveform data growing faster than model size as compute increases.
\ding{183} When waveform data are limited, increase FM size to
improve data efficiency and clinical performance.
Unlike web-scale language corpora, real-world clinical waveform datasets
are difficult to expand continuously because of patient privacy, legal
restrictions, and institutional access
constraints~\citep{johnson2016mimic,rieke2020federated}.
At fixed $D$ and $P$, larger FMs use the available waveforms more
efficiently and continue to improve~clinical~performance.

\begin{figure}[!ht]
\centering
\includegraphics[width=\linewidth]{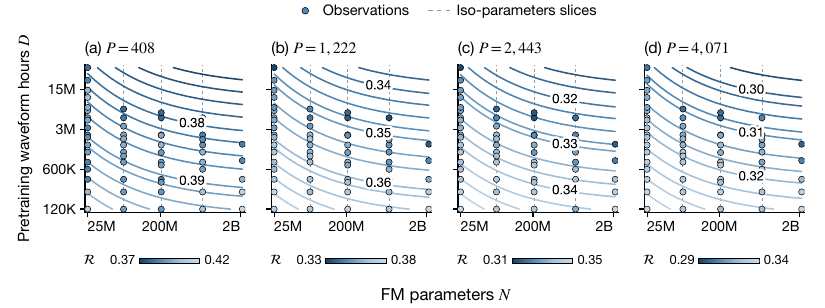}
\caption{Iso-error contours for FM size and pretraining waveform hours.
Each curve holds the predicted eight-task average ranking error fixed
under Equation~\ref{eq:ndp-scaling-law}.
(a--d) $P=408$, $1{,}222$, $2{,}443$, and $4{,}071$, respectively.
Points show measurements, with color indicating observed ranking error.
Dashed lines mark iso-parameters slices.
Patient counts are averaged across the eight tasks.
Moving right along a curve reduces waveform hours at fixed ranking error $\risk$.}
\label{fig:waveform-data-efficiency}
\end{figure}

\paragraph{Q2: How much waveform data can larger FMs save?}
Larger FMs use pretraining waveform data more efficiently.
At fixed $P$ and target ranking error,
Equation~\ref{eq:ndp-scaling-law} gives $AN^{-a}+BD^{-b}=\mathrm{constant}$.
The iso-error contours in Figure~\ref{fig:waveform-data-efficiency} trace
model--data tradeoffs across all four labeled-patient cohorts: increasing $N$ reduces
the waveform hours $D$ needed for the same predicted performance.
The saving varies with the starting $N$ and $D$.

We draw the following implications:
\ding{182} More waveform data increases the benefit of model scaling.
For example, at $P=4{,}071$, scaling from 20M to 720M yields a relative
reduction in predicted ranking error of $4.0\%$ at 36.3M pretraining
waveform hours, compared with $2.6\%$ at 120K hours.
\ding{183} Scaling pretraining data yields larger relative gains with
more labeled patients. For a 720M FM, increasing $D$ from 120K to 36.3M
hours yields a relative reduction in predicted ranking error of $10.0\%$
at $P=4{,}071$, compared with $6.3\%$ at $P=408$.
These complementary gains support scaling FM capacity together with
pretraining waveform data.

\paragraph{Q3: How can scaling pretraining reduce labeling requirements?}
Scaling pretraining reduces the labeled-patient count needed to attain
a given clinical performance. For a target ranking error $\risk_{\mathrm{t}}$,
solving Equation~\ref{eq:ndp-scaling-law} for $P$ gives
\begin{equation}
P=\left[\left(AN^{-a}+BD^{-b}\right)
\left(\risk_{\mathrm{t}}^{-1}-2\right)\right]^{1/g}.
\label{eq:required-patients}
\end{equation}
\begin{figure}[tbp]
\centering
\begin{minipage}{0.62\linewidth}
\centering
\includegraphics[width=\linewidth]{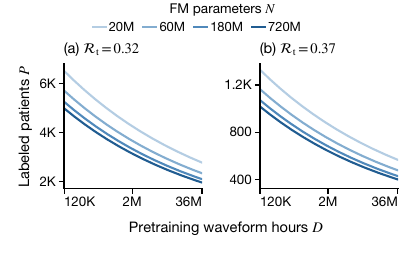}
\setlength{\abovecaptionskip}{3pt}
\caption{Iso-ranking-error curves showing predicted labeling requirements
averaged across eight tasks. Target ranking error $\risk_{\mathrm{t}}$
is 0.32 in (a), 0.37 in (b).}
\label{fig:labeling-requirements}
\end{minipage}
\end{figure}
Figure~\ref{fig:labeling-requirements} translates pretraining resources
into labeling requirements at two target errors across all five FM sizes.
Increasing either $N$ or $D$ consistently reduces the required $P$.

We draw the following implications:
\ding{182} Model scaling improves label efficiency.
At $\risk_{\mathrm{t}}=0.32$ and 4.8M pretraining waveform hours, scaling
from 20M to 720M reduces the predicted patient requirement by $27.2\%$,
from approximately 3,733 to 2,717.
\ding{183} Longer pretraining further reduces annotation requirements.
For the 720M FM at the same target error, increasing $D$ from 120K to
36.3M hours reduces the requirement by $61\%$, from approximately
4,986 to 1,949.

\FloatBarrier
% \paragraph{Q4: Does scaling improve prediction at a fixed alert budget?}

\section{Conclusion}
We establish a predictive scaling law for \aether{} through linear probing
on eight clinical tasks, with generalization to larger FMs and longer
pretraining runs. Our study yields three findings.
(1)~Compute-optimal training scales FM size and pretraining hours together,
with hours growing faster. The fitted law allocates a $10\times$ compute
increase to $1.2\times$ larger FMs and $8.2\times$ more pretraining hours.
(2)~Larger FMs use waveform data more efficiently, and more pretraining data
amplifies model-scaling gains. Doubling a 25M FM from a 4.8M-hour reference
reduces predicted waveform requirements by $51.8\%$ at fixed performance
and labeled-patient count.
(3)~Pretraining and clinical supervision reinforce each other: more labeled
patients increase pretraining gains, while larger FMs and longer pretraining
reduce labeling requirements. Extending 720M pretraining from 120K to
36.3M hours reduces the predicted patient requirement by $61\%$ at a
target ranking error of $0.32$. These findings provide a quantitative
recipe for using additional compute to reach target clinical performance
with fewer waveform hours and labeled patients. Scaling our training recipe offers a promising and
durable path toward physiological modeling and clinical prediction.

\FloatBarrier
\Needspace{8\baselineskip}
\bibliographystyle{style/icml2025}
\bibliography{999_reference}

\clearpage
\appendix
\section{Clinical downstream datasets}
\label{app:clinical-downstream-datasets}

We construct eight clinical prediction tasks by linking physiological waveforms
to clinical records in MIMIC-III. Each task uses one hour of waveforms to
predict a future clinical outcome. Below, we describe the data sources,
prediction windows, task definitions, and eligibility criteria.
Table~\ref{tab:cohorts} reports cohort sizes for all three data splits.

\subsection{Data sources and patient splits}
\label{app:downstream-sources}
We use the MIMIC-III Waveform Database Matched
Subset~\citep{moody2020mimicwaveform}, which contains bedside recordings
of ECG, arterial blood pressure, plethysmography, and respiration.
Waveforms are sampled at 125\,Hz, with available channels varying across
recordings. MIMIC-III v1.4~\citep{johnson2016mimic} provides vital signs,
laboratory results, medications, procedures, ventilator settings, and outcomes.
Eligible ICU stays involve adults with at least one matched waveform record.
All downstream patients are held out from pretraining. We split patients
into disjoint training, validation, and test cohorts before constructing
windows, keeping all stays from each patient together.

\subsection{Prediction windows}
\label{app:downstream-geometry}
Each example has a prediction time $T$, a clinical history
$[T-W,\,T)$, a forecast gap $[T,\,T+G)$, and an outcome horizon
$[T+G,\,T+G+H)$. We use six hours of clinical history ($W=6$\,h)
to determine eligibility and prior clinical status. The FM receives only
the one-hour waveform segment ending at $T$; the ``Hours'' columns in
Table~\ref{tab:cohorts} sum these input durations.
Prediction times follow a sliding schedule for surveillance tasks,
daily checkpoints for ventilator readiness, and documented treatment
events for fluid response.
Table~\ref{tab:downstream-geometry} summarizes these task-specific settings.

\begin{table}[ht]
\centering\footnotesize
\setlength{\tabcolsep}{4pt}
\caption{Prediction windows and eligibility for the eight downstream tasks.
$W$ denotes clinical history, $G$ the forecast gap, and $H$ the outcome horizon.
Each FM input contains one hour of waveforms ending at $T$.
``Targets'' gives the number of binary outcomes evaluated per task.}
\label{tab:downstream-geometry}
\resizebox{\linewidth}{!}{%
\begin{tabular}{@{}llccccl@{}}
\toprule
\midrule
Task & Prediction time & $W$ & $G$ & $H$ & Targets & Eligibility at $T$ \\
\midrule
Acute deterioration & Sliding & 6\,h & 0 & 6\,h & 3 & Eligible for at least one component \\
Circulatory deterioration & Sliding & 6\,h & 0 & 6\,h & 1 & No active vasopressors \\
Dynamic mortality & Sliding & 6\,h & 6\,h & 24\,h & 1 & All monitored patients \\
Respiratory escalation & Sliding & 6\,h & 0 & 6\,h & 1 & No invasive ventilation \\
Sepsis onset & Sliding & 6\,h & 1\,h & 5\,h & 1 & No prior Sepsis-3 onset \\
Fluid response & Infusion start & 6\,h & 30\,min & 30\,min & 2 & Documented rapid crystalloid course \\
KDIGO 2/3 AKI & Sliding & 6\,h & 1\,h & 23\,h & 2 & No stage 2/3 AKI or maintenance dialysis \\
Ventilator readiness & Daily checkpoint & 6\,h & 0 & 24\,h & 1 & Active invasive ventilation \\
\midrule
\bottomrule
\end{tabular}%
}
\end{table}

\subsection{Task definitions}
\label{app:downstream-tasks}

\noindent\textbf{Acute deterioration.}
We predict three binary outcomes within six hours: initiation of vasopressors,
initiation of invasive mechanical ventilation, and a composite of either event.
The first two targets include patients who are not receiving the corresponding
treatment at $T$. The composite includes patients eligible for at least one
component and is positive when an eligible component occurs.

\noindent\textbf{Circulatory deterioration.}
We predict the first new vasopressor initiation within six hours among
patients receiving no vasopressors at $T$.

\noindent\textbf{Dynamic mortality.}
We predict in-hospital death during $[T+6\,\mathrm{h},\,T+30\,\mathrm{h})$.
All monitored patients are eligible at $T$, with a six-hour gap between
the waveform input and the outcome horizon.

\noindent\textbf{Respiratory escalation.}
We predict new invasive ventilation within six hours among patients
receiving no support, oxygen, high-flow nasal cannula, or non-invasive
ventilation at $T$.

\Needspace{5\baselineskip}
\noindent\textbf{Sepsis onset.}
We predict the first Sepsis-3 onset during $[T+1\,\mathrm{h},\,T+6\,\mathrm{h})$.
Onset requires suspected infection, identified by a culture order paired
with antibiotic administration, and an acute increase of at least two points
in the Sequential Organ Failure Assessment (SOFA) score.
Patients with sepsis onset at or before $T$ are ineligible.

\noindent\textbf{Fluid response.}
The prediction time $T$ is the start of a documented rapid crystalloid course;
overlapping courses are merged into a single event.
We compare median mean arterial pressure during minutes 30--60 after
treatment begins with the pre-treatment reference.
The two targets require increases of at least 15\% and 5\,mmHg, respectively.
These endpoints measure the documented blood-pressure response following fluid administration.

\noindent\textbf{KDIGO stage 2/3 acute kidney injury.}
We predict the first creatinine-defined stage 2 or 3 acute kidney injury
during $[T+1\,\mathrm{h},\,T+24\,\mathrm{h})$, using Kidney Disease:
Improving Global Outcomes (KDIGO) criteria.
Eligible patients have no prevalent stage 2/3 injury or maintenance dialysis
and at least one creatinine measurement in the outcome horizon.
A second target uses urine-output criteria when hourly urine measurements,
a reliable body weight, and catheter information are available.

\noindent\textbf{Ventilator readiness.}
At each daily checkpoint during invasive ventilation, we predict liberation
within 24 hours. The label is positive if ventilation ends during this
period and the patient remains free of invasive ventilation for the
following 48 hours.

\subsection{Label availability and eligibility}
\label{app:downstream-labels}
We assign labels only when the outcome can be determined from the available
records. For incident-event targets, patients with the event already present
at $T$ are ineligible. Surveillance examples with an event during the forecast
gap are also excluded. A negative label requires sufficient follow-up to
establish that the event did not occur within the outcome horizon.
Unavailable labels are masked from the corresponding target's loss and metrics.
Eligibility and label availability are determined separately for each target.
Following Section~\ref{sec:linear-probing}, we compute patient-balanced AUROC
for each target and average across targets to obtain the task score.
Repeated or overlapping windows retain their patient identity; each patient's
inputs receive equal total weight within a target.

\subsection{Leakage controls}
\label{app:downstream-leakage}
FM inputs contain only waveform samples recorded before $T$.
Clinical history used to determine eligibility is restricted to information
charted before $T$; subsequent records are used only to establish outcomes.
Post-prediction interventions, final diagnoses and disposition, realized
length of stay, and outcome labels are excluded from model inputs.
Documented treatment before $T$ informs eligibility: for example, active
vasopressor treatment excludes a new-start target.
Waveform normalization uses only the input window, and feature standardization
for linear probing uses the selected training patients.
For each example, we retain its data version, patient split, prediction time,
input and outcome intervals, and available-channel mask to support reproduction.

\clearpage
\section{Expected Forms of the Scaling Law}
\label{app:expected-forms}

\paragraph{Ranking risk.}
For patient-weighted positive and negative examples $X^+$ and $X^-$,
the ranking error of a score $s$ is~\citep{agarwal2014ranking}:
\begin{equation}
\risk(s)=\Pr\!\left[s(X^+)<s(X^-)\right]
+\frac12\Pr\!\left[s(X^+)=s(X^-)\right]
=1-\operatorname{AUROC}(s).
\label{eq:ranking-risk-definition}
\end{equation}
A constant score gives $\risk=1/2$.
Macro averaging applies this definition across targets and tasks,
yielding the eight-task ranking error modeled in
Section~\ref{sec:scaling-law}.

\paragraph{Resource-dependent contributions.}
Following the expected-risk decomposition of
\citet{hoffmann2022chinchilla}, let $\risk_*$ be the best population
ranking risk and $\risk_N^*$ the best risk attainable by a size-$N$ FM
and a linear probe. For the pretrained FM representation
$\widehat\phi_{N,D}$, define
$\risk_{\rm LP}^*(\widehat\phi_{N,D})$
as the best population risk of a linear probe on that representation.
With the learned score $\widehat s_{N,D,P}$, we obtain
\begin{align}
\mathbb E\risk(\widehat s_{N,D,P})
={}&\risk_*+\left(\risk_N^*-\risk_*\right)\nonumber\\
&+\left[\mathbb E\risk_{\rm LP}^*(\widehat\phi_{N,D})-\risk_N^*\right]
\nonumber\\
&+\left[\mathbb E\risk(\widehat s_{N,D,P})
-\mathbb E\risk_{\rm LP}^*(\widehat\phi_{N,D})\right].
\label{eq:ranking-risk-decomposition}
\end{align}
The three resource-dependent terms describe the contributions of FM
capacity, pretraining, and labeled-patient supervision. For the
theoretical data-scaling analysis, we adopt the single-pass view of
\citet{hoffmann2022chinchilla}: $D$ measures pretraining data and the
associated training progress. Frozen linear probing adds the
supervision resource $P$ to this description.

\paragraph{Expected forms of the scaling terms.}
To model the observed scaling regime, we express ranking error relative
to its advantage over chance through $\risk/(1-2\risk)$ for
$0<\risk<1/2$. Motivated by the capacity and training contributions in
Equation~\ref{eq:ranking-risk-decomposition}, we assume decreasing
power-law terms $AN^{-a}$ and $BD^{-b}$, with positive amplitudes and
exponents. We model the response to labeled-patient supervision with a
common factor $P^{-g}$. These assumptions give
\begin{equation}
\frac{\widehat\risk}{1-2\widehat\risk}
=(AN^{-a}+BD^{-b})P^{-g}.
\label{eq:expected-ranking-terms}
\end{equation}
Solving directly for predicted ranking error yields
\begin{equation}
\widehat\risk(N,D,P)
=\left[2+\frac{P^g}{AN^{-a}+BD^{-b}}\right]^{-1}.
\label{eq:expected-ranking-law}
\end{equation}
The resulting response lies between zero and the chance error of $1/2$
and decreases with all three resources. It also preserves the allocation
ordering implied by the power-law terms, connecting pretraining and
supervision in a common predictive model.

\paragraph{Fitting the parametric law.}
We fit $A,B,a,b,g$ by nonlinear least squares on $K$ complete
eight-task resource configurations\footnote{We use $K=140$ resource
configurations in our experiments.}:
\begin{equation}
\min_{A,B,a,b,g>0}\ \sum_{i=1}^{K}
\left(\widehat\risk(N_i,D_i,P_i)-\risk_i\right)^2.
\label{eq:ranking-law-fit}
\end{equation}
Each outcome is the equal-task mean of patient-level macro AUROC
converted to ranking error. The fit uses exact FM parameter counts,
pretraining waveform hours, and the corresponding mean absolute
labeled-patient counts. We use 30 deterministic initializations and
retain the solution with the smallest objective. Full-precision
coefficients generate every prediction and contour; displayed
coefficients are rounded to three decimal places.
For resource-level prediction, we omit all configurations at one value
of $N$, $D$, or $P$ and refit on the remaining configurations. There are
four model-size levels, six pretraining-hour levels, and four
patient-cohort levels. We average prediction MAE across levels within
each resource axis and then equally across the three axes.
The 2.1B and longer-duration predictions use the same fitted law,
connecting this procedure to Section~\ref{sec:scale-extrapolation}.

\paragraph{Resource gains and data efficiency.}
The fitted law makes proportional resource gains explicit:
\begin{align}
-\frac{\partial\log\widehat\risk}{\partial\log N}
&=(1-2\widehat\risk)\frac{aAN^{-a}}{AN^{-a}+BD^{-b}},\\
-\frac{\partial\log\widehat\risk}{\partial\log D}
&=(1-2\widehat\risk)\frac{bBD^{-b}}{AN^{-a}+BD^{-b}},\\
-\frac{\partial\log\widehat\risk}{\partial\log P}
&=(1-2\widehat\risk)g.
\label{eq:ranking-resource-elasticity}
\end{align}
All three quantities are positive. The labeled-patient response is
largest throughout the fitting grid, supporting its role as the
leading downstream scaling factor.
At fixed $P$ and target error, changing \mbox{FM size from $N$ to $N'$} gives
\begin{equation}
D'=\left[D^{-b}+\frac{A}{B}\left(N^{-a}-{N'}^{-a}\right)\right]^{-1/b}.
\label{eq:ranking-data-efficiency}
\end{equation}
Thus a larger FM reaches the same predicted performance using fewer
pretraining waveform hours. The example in
Section~\ref{sec:scaling-law} uses $N=25$M, $D=4.8$M hours, and
$N'=50$M, giving $D'=2.3$M hours with the full-precision fit.

\paragraph{Compute-optimal allocation.}
For fixed $P$, Equation~\ref{eq:expected-ranking-law} increases with
$AN^{-a}+BD^{-b}$. Under $C\simeq6qND$, its minimum therefore solves
\begin{equation}
\min_N\left[AN^{-a}+B\left(\frac{C}{6qN}\right)^{-b}\right].
\end{equation}
Differentiating with respect to $\log N$ balances the two marginal
contributions, $aAN^{-a}=bBD^{-b}$, and gives
Equations~\ref{eq:compute-size-optimum}--\ref{eq:compute-hours-scaling}.
At each compute value, the green frontier in
Figure~\ref{fig:compute-efficient-frontier} uses $D=C/(6qN)$ and
minimizes error over the observed FM-size range. The minimum and maximum
FM sizes bound the allocation, giving a frontier with interior and
boundary solutions.

\end{document}